\documentclass[conference]{IEEEtran}
\usepackage{stmaryrd}
\usepackage{amsfonts}

\usepackage{graphicx}
\usepackage{tabularx}
\usepackage{colortbl}
\usepackage{setspace}

\usepackage{url}
\usepackage{amsmath}
\usepackage{amssymb}
\usepackage{array}
\usepackage{longtable}

\IEEEoverridecommandlockouts    

\begin{document}

\title{\LARGE\bf Infrared Face Recognition: A Literature Review}

\author{Reza Shoja Ghiass$^\dag$, Ognjen Arandjelovi\'c$^\ddag$, Hakim Bendada$^\dag$, and Xavier Maldague$^\dag$\\$^\dag$~Universit\'{e} Laval, Quebec, Canada ~~~~~~~~~~ $^\ddag$~Deakin University, Geelong, Australia}


\maketitle

\begin{abstract}
Automatic face recognition (AFR) is an area with immense practical potential which includes a wide range of commercial and law enforcement applications, and it continues to be one of the most active research areas of computer vision. Even after over three decades of intense research, the state-of-the-art in AFR continues to  improve, benefiting from advances in a range of different fields including image processing, pattern recognition, computer graphics and physiology. However, systems based on visible spectrum images continue to face challenges in the presence of illumination, pose and expression changes, as well as facial disguises, all of which can significantly decrease their accuracy. Amongst various approaches which have been proposed in an attempt to overcome these limitations, the use of infrared (IR) imaging has emerged as a particularly promising research direction. This paper presents a comprehensive and timely review of the literature on this subject.
\end{abstract}


\section{Introduction}\label{s:intro}
\PARstart{I}{n} the last two decades AFR has consistently been one of the most active research areas of computer vision and applied pattern recognition. Systems based on images acquired in the visible spectrum have reached a significant level of maturity with some practical success. However, a range of nuisance factors continue to pose serious problems when visible spectrum based AFR methods are applied in a real-world setting. Dealing with illumination, pose and facial expression changes, and facial disguises is still a major challenge.

There is a large corpus of published work which has attempted to overcome the aforesaid difficulties by developing increasingly sophisticated models which were then applied on the same type of data -- usually images acquired in the visible spectrum (wavelength approximately in the range $390-750$~nm). Pose, for example, has been normalized by a learnt 2D warp of an input image \cite{GrosMattBake2006}, generated from a model fitted using an analysis-by-synthesis approach \cite{BlanVett2003} or synthesized using a statistical method \cite{MogaPrinKaut2009}, while illumination has been corrected for using image processing filters \cite{Aran2013,Aran2012e,NishYama2006} and statistical facial models \cite{WolfShas2003}, amongst others, with varying levels of success. Other methods adopt a multi-image approach by matching sets \cite{AranCipo2006e} or sequences of images \cite{BowyChanFlynChen2006}. Another increasingly active research direction has pursued the use of alternative modalities. For example, it is clear that data acquired using 3D scanners \cite{PanWu2005} is inherently robust to illumination and pose changes. However, the cost of these systems is high and the process of data collection overly restrictive.

\subsection{Infrared Spectrum} 
\begin{figure*}[htb]
  \centering
  \includegraphics[width=1.00\textwidth]{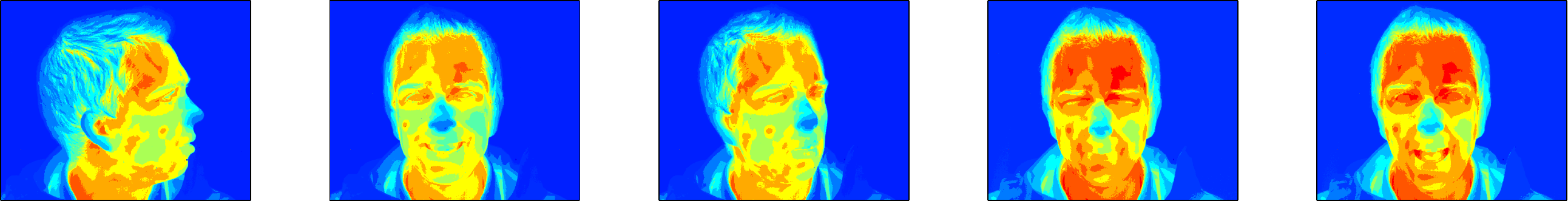}
  \caption{ False colour thermal appearance images of a subject in five arbitrary poses and facial expressions from the \textit{Thermal IR Face Motion} data set
            acquired by the authors and freely available upon request. }
  \label{f:dbFM}
\end{figure*}

IR imagery is a modality which has attracted particular attention, in large part due to its invariance to the changes in illumination by visible light; a few examples of thermal IR images are shown in Fig.~\ref{f:dbFM}. A detailed account of the relevant physics, which is outside the scope of this paper, can be found in \cite{Mald2001}. In the context of AFR, data acquired using IR cameras has distinct advantages over the more common cameras which operate in the visible spectrum. For instance, IR images of faces can be obtained under any lighting condition, even in completely dark environments, and there is some evidence that IR ``appearance'' may exhibit a higher degree of robustness to facial expression changes~\cite{FrieYesh2003}. IR energy is also less affected by scattering and absorption by smoke or dust than visible light \cite{NicoSchm2011}. Unlike visible spectrum imaging, IR imaging can be used to extract not only exterior, but also useful subcutaneous anatomical information, such as the vascular network of a face~\cite{BuddPavlTsiaBaza2007}. Lastly, in contrast to visible spectrum imaging, thermal vision can be used to detect facial disguises \cite{PavlSymo2000}.

\subsection{Spectral Composition}In the literature, it has been customary to divide the IR spectrum into four sub-bands: near IR (NIR; wavelength $0.75-1.4\mu$m), short wave IR (SWIR; wavelength $1.4-3\mu$m), medium wave IR (MWIR; wavelength $3-8\mu$m), and long wave IR (LWIR; wavelength $8-15\mu$m).
This division of the IR spectrum is also observed in the manufacturing of IR cameras, which are often made with sensors that respond to electromagnetic radiation constrained to a particular sub-band.

It should be emphasized that the division of the IR spectrum is not arbitrary. Rather, different sub-bands correspond to continuous frequency chunks of the solar spectrum which are divided by absorption lines of different atmospheric gasses \cite{Mald2001}. In the context of AFR, one of the largest differences between different IR sub-bands emerges as a consequence of the human body's heat emission spectrum. Specifically, most of the heat energy is emitted in LWIR sub-band, which is why it is often referred to as the thermal sub-band (this term is sometimes extended to include the MWIR sub-band). Significant heat is also emitted in the MWIR sub-band. Both of these sub-bands can be used to \emph{passively} sense facial thermal emissions without an external source of light. This is one of the reasons why LWIR and MWIR sub-bands have received the most attention in the AFR literature. In contrast to them, facial heat emission in the SWIR and NIR sub-bands is very small and AFR systems operating on data acquired in these sub-bands require appropriate illuminators i.e.\ recognition is \emph{active} in nature. In recent years, the use of NIR also started received increasing attention from the AFR community, while the utility of the SWIR sub-band has yet to be studied in depth.

\subsection{Challenges} The use of IR images for AFR is not void of its problems and challenges. For example, MWIR and LWIR images are sensitive to the environmental temperature, as well as the emotional, physical and health condition of the subject. They are also affected by alcohol intake. Another potential problem is that eyeglasses are opaque to the greater part of the IR spectrum (LWIR, MWIR and SWIR) \cite{TasmJaeg2009}. This means that a large portion of the face wearing eyeglasses may be occluded, causing the loss of important discriminative information. Unsurprisingly, each of the aforementioned challenges has led to and motivated a new research direction. Some researchers have suggested fusing the information from IR and visible modalities as a possible solution to the problem posed by the opaqueness of eyeglasses \cite{KongHeoAbidPaik+2005}. Others have described methods which use IR images to extract a range of invariant features such as facial vascular networks \cite{BuddPavlTsiaBaza2007} or blood perfusion data \cite{WuWeiFangLi+2007} in order to overcome the temperature dependency of thermal appearance.

Another consideration of interest pertains to the impact of sunlight if recognition is performed outdoors and during daytime.  Although invariant to the changes in the illumination by visible light itself (by definition), the IR ``appearance'' in the NIR and SWIR sub-bands \emph{is} affected by sunlight which has significant spectral components at the corresponding wavelengths. This is one of the key reasons why NIR and SWIR based systems which perform well indoors struggle when applied outdoors \cite{LiChuLiaoZhan2007}.

\section{IR: Advantages and Disadvantages in AFR}\label{s:ir}
Many of the methods for IR based AFR have been inspired by or are verbatim copies of algorithms which were initially developed for visible spectrum recognition. In most cases, these methods make little use of the information about the spectrum which was used to acquire images. However, the increasing appreciation of challenges encountered in trying to robustly match IR images, strongly suggests that domain specific properties of data should be exploited more. Indeed, as we will discuss in Sections~\ref{s:main} and~\ref{s:summary}, the recent trend in the field has been moving in this direction, increasingly complex IR specific models being proposed. Thus, in this section we focus on the relevant differences of practical significance between IR and visible spectrum images. The use of IR imagery provides several important advantages as well as disadvantages, and we start with a summary of the former.


\subsection{Advantages of Infrared Data in AFR}
Much of the early work on the potential of IR images as identity signatures was performed by Prokoski \textit{et al.}\ \cite{ProkRiedCoff1992}. They were the first to advance the idea that IR ``appearance'' could be used to extract robust biometric features which exhibit a high degree of uniqueness and repeatability.

Facial expression and pose changes are two key factors that an AFR system should be robust to for it to be useful in most practical applications of interest. By comparing image space differences of thermal and visible spectrum images, Friedrich \textit{et al.}\ \cite{FrieYesh2003} found that thermal images are less affected by changes in pose or facial expression than their visible spectrum counterparts. Illumination invariance of different IR sub-bands was analyzed in detail by Wolff \textit{et al.}\ \cite{WolfSocoEvel2001} who showed the superiority of IR over visible data with respect to this important nuisance variable.

The very nature of thermal imaging also opens the possibility of non-invasive extraction and use of superficial anatomical information for recognition. Blood vessel patterns are one such example. As they continually transport circulating blood, blood vessels are somewhat warmer than the surrounding tissues. Since thermal cameras capture the heat emitted by a face, standard image processing techniques can be readily used to extract blood vessel patterns from facial thermograms. An important property of these patterns which makes them particularly attractive for use in recognition is that the blood vessels are ``hardwired'' at birth and form a pattern which remains virtually unaffected by factors such as aging, except for predictable growth \cite{PersBusc2011}. Moreover, it appears that the human vessel pattern is robust enough to facilitate scaling up to large populations \cite{ProkRiedCoff1992}. Prokoski \textit{et al.}\ estimate that about 175 blood vessel based minutiae can be extracted from a full facial image which, they argued, can exhibit a far greater number of possible configurations than the size of the foreseeable maximum human population. It should be noted that the authors did not propose a specific algorithm to extract the minutiae in question. In the same work, the authors also argued that forgery attempts and disguises can both be detected by IR imaging. The key observation is that the temperature distribution of artificial facial hair or other facial wear differs from that of natural hair and skin, allowing them to be differentiated one from another.

\subsubsection{The Twin Paradox}
An interesting question first raised by Prokoski \textit{et al.}\ \cite{ProkRied1998} concerns thermograms of monozygotic twins. The appearance of monozygotic twins (or ``identical'' in common vernacular) is nearly identical in the visible spectrum. Using a small number of thermograms of monozygotic twins which were qualitatively assessed for similarity, Prokoski \textit{et al.}\ found that the difference in appearance was significantly greater in the thermal than in the visible spectrum, and sufficiently so to allow for them to be automatically differentiated. This hypothesis was disputed by subsequent contradictory findings of Chen \textit{et al.}\ \cite{ChenFlynBowy2005}. However, the weight of evidence provided both by Prokoski \textit{et al.}\ as well as Chen \textit{et al.}\ is inadequate to allow for a confident conclusion to be made. Both positive and negative claims are based on experiments which use little data and lack sufficient rigour. In addition, it is plausible that the truth may be somewhere in the middle, that is, that in some cases monozygotic twins can be differentiated from their thermograms and in others not, depending on a host of physiological variables.

\subsection{Limitations of Infrared Data in AFR}
In the context of AFR, the main drawback specific to the thermal sub-band images (or \emph{thermograms}, as they are often referred to), the most often used sub-band of the IR spectrum, stems from the fact that the heat pattern emitted by the face is affected by a number of confounding variables, such as ambient temperature, air flow conditions, exercise, postprandial metabolism, illness and drugs \cite{ProkRied1998}. Some of the confounding variables produce global, others local thermal appearance changes. Wearing clothes, experiencing stress, blushing, having a headache or an infected tooth are examples of factors which can effect localized changes.

The high sensitivity of the facial thermogram to a large number of extrinsic factors makes the task of finding persistent and discriminative features a challenging one. It also lends support to the ideas first voiced by Prokoski \textit{et al.}\ who argued against the use of thermal appearance based methods in favour of anatomical feature based approaches invariant to many of the aforementioned factors. As we will discuss in Sec.~\ref{ss:features}, this direction of IR based AFR has indeed attracted a substantial research effort.

Another drawback of using the IR spectrum for AFR is that glass and thus eyeglasses are opaque to wavelengths longer and including the SWIR sub-band. Consequently an important part of the face, one rich in discriminative information, may be occluded in the corresponding images. In particular, the absence of appearance information around the eyes can greatly decrease recognition accuracy~\cite{AranHammCipo2010}. Multi-modal fusion based methods have been particularly successful in dealing with this problem, as described in detail in Sec.~\ref{sss:glasses}.

Lastly, a major challenge when NIR and SWIR sub-bands are used for recognition, stems from their sensitivity to sunlight which has significant spectral components at the corresponding wavelengths \cite{LiChuLiaoZhan2007}. In this sense, the problem of matching images acquired in NIR and SWIR sub-bands is similar to matching visible spectrum images.

\section{Face Recognition Using Infrared}\label{s:main}
In this review, we recognize four main groups of AFR methodologies which use IR data: holistic appearance based, feature based, multi-spectral based, and multi-modal fusion based. Holistic appearance methods use the entire IR appearance image of a face for recognition. Feature based approaches use IR images to extract salient face features, such as facial geometry, its vascular network or blood perfusion data. Spectral model based approaches model the process of IR image formation to decompose images of faces. Some approaches directly use data from multi-spectral or hyper-spectral imaging sensors to obtain facial images across different frequency sub-bands. Multi-modal fusion based approaches combine information contained in IR images with information contained in other types of modalities, such as visible spectrum data, with the aim of exploiting their complementary advantages. As the understanding of the challenges of using IR data for AFR has increased, this direction of research has become increasingly active.

\subsection{Appearance-Based Methods}\label{ss:appearance}
The earliest attempts at examining the potential of IR imaging for AFR dates back to 1992 and the work done by Prokoski \textit{et al.}\ \cite{ProkRiedCoff1992}. Their work introduced the concept of ``elementary shapes'' extracted from thermograms, which are likened to fingerprints. While precise technical detail of the method used to extract these elementary shapes is lacking, it appears that they are isothermal regions segmented out from an image. There is no published record on the effectiveness of this representation.

\subsubsection{Early Approaches}\label{ss:early}
Perhaps unsurprisingly, most of the automatic methods which followed the work of  Prokoski \textit{et al.}\ closely mirrored in their approach methods developed for the more popular visible spectrum based recognition. Generally, these used holistic face appearance in a simple statistical manner, with little attempt to achieve any generalization, relying instead on the availability of training data with sufficient variability of possible appearance for each subject.

One of the first attempts at using IR data in an AFR system was described by Cutler \cite{Cutl1996}. His method was entirely based on the popular Eigenfaces method proposed by Turk and Pentland \cite{TurkPent1991}. Using a database of 288 thermal images (12 images for each of the 24 subjects in the database) which included limited pose and facial expression variation, Cutler reported rank-1 recognition rates of 96\% for frontal and semi-profile views, and 100\% for profile views. These recognition rates compared favourably with those achievable using the same methodology on visible spectrum images. Following these promising results, many of the subsequently developed algorithms also adopted Eigenfaces as the baseline classifier. For example, findings similar to those made by Cutler were independently reported by Socolinsky \textit{et al.}\ \cite{SocoWolfNeuhEvel2001}.

In their later work, Socolinsky and Selinger \textit{et al.}\ \cite{SocoSeli2004} extended their comparative evaluation of thermal and visible data based recognition using a wider range of linear methods: Eigenfaces (that is, principal component analysis), linear discriminant analysis, local feature analysis and independent component analysis. Their results corroborated previous observations made in the literature on the superiority of the thermal spectrum for recognition in the presence of a range of nuisance variables. However, the conclusions that could be drawn from their comparative analysis of different recognition approaches or indeed that of Culter, were limited by the insufficiently challenging data sets which were used: pose and expression variability was small, training and test data were acquired in a single session, and the subjects wore no eyeglasses. This is reflected in the fact that all of the evaluated algorithms achieved comparable, and in practical terms high, recognition rates (approximately 93-98\% on average).

\subsubsection{Effects of Registration}\label{ss:registration}
In practice, after face detection, faces are still insufficiently well aligned (registered) for pixel-wise comparison to be meaningful. The simplest and the most direct way of registering faces is by detecting a discrete set of salient facial features and then applying a geometric warp to map them into a canonical frame. Unlike in the case of images acquired in the visible spectrum, in which several salient facial features (such as the eyes and the mouth) can usually be reliably detected \cite{ChriCootScot2004,AranCipo2004a,DingMart2010}, most of the work to date supports the conclusion that salient facial feature localization in thermal images is significantly more challenging. Different approaches, which mainly focus on the eyes, were described by Tzeng \textit{et al.}\ \cite{TzenLeeChen2011}, Arandjelovi\'c \textit{et al.}\ \cite{AranHammCipo2010}, Jin \textit{et al.}\ \cite{JinShouXiuGu2009}, Bourlai \textit{et al.}\ \cite{BourWhitKaka2011} and Martinez \textit{et al.}\ \cite{MartBinePant2010}. What is more, the effect of feature localization errors and thus registration errors seems to be greater for thermal than visible spectrum images. This was investigated by Chen \textit{et al.}\ \cite{ChenFlynBowy2003} who demonstrated a substantial reduction in thermal based recognition rates when small localization errors were synthetically introduced to manually marked eye positions.

Zhao \textit{et al.}\ \cite{ZhaoGrig2005} ingeniously circumvent the problem of localizing the eyes in passively acquired images by their use of additional active NIR data. A NIR lighting source placed close to and aligned with the camera axis is used to illuminate the face. Because the interior of the eyes reflects the incident light the pupils appear distinctively bright and as such are readily detected in the observed image (the so-called ``bright pupil'' effect). Zhao \textit{et al.}\ use the locations of pupils to register images of faces, which are then represented using their DCT coefficients and classified using a support vector machine.

\subsubsection{Recent Advances in IR Appearance Based Recognition}\label{ss:modern}
Although the general trend in the field has been way from appearance based approaches and in the direction of feature and model based methods, the former have continued to attract some research interest. Much like the initial work, the recent advances in appearance based IR AFR has closely mirrored research in visible spectrum based recognition. Progress in comparison with the early work is mainly to be found in the use of more sophisticated statistical techniques. For example, Elguebaly and Bouguila \cite{ElguBoug2011} recently described a method based on a generalized Gaussian mixture model, the parameters of which are learnt from a training image set using a Bayesian approach. Although substantially more complex, this approach did not demonstrate a statistically significant improvement in recognition on the IRIS Thermal/Visible database, both methods achieving rank-1 rate of approximately 95\%. Lin \textit{et al.}\ \cite{LinWenrLiZhij2011} were the first to investigate the potential of the increasingly popular compressive sensing in the context of IR AFR. Using a proprietary database of 50 persons with 10 images each person, their results provided some preliminary evidence for the superiority of this approach over wavelet based decomposition.

Considering that the development of appearance based methods has nearly exclusively focused on the use of more sophisticated statistical techniques (rather than the incorporation of data specific knowledge, say), it is a major flaw in this body of research that the data sets used for evaluation have not included the types of intra-personal variations that appearance based methods are likely to be sensitive to. Indeed, none of the data sets that we are aware of included intra-personal variations due to differing emotional states, alcohol intake or exercise, for example, or even ambient temperature. This observation casts a shadow on the reported results and impedes further development of algorithms which could cope with such variations in a realistic, practical setup.

\subsection{Feature-Based Methods}\label{ss:features}
An early approach which uses features extracted from thermal images, rather than raw thermal appearance, was proposed by Yoshitomi \textit{et al.}\ \cite{YoshMiyaTomiKimu1997}. Following the localization of a face in an image, their method was based on combining the results of neural network based classification of grey level histograms and locally averaged appearance, and supervised classification of a facial geometry based descriptor. The proposed method was evaluated across room temperature variations ranging from 302K to 285K. As expected, the highest recognition rates were attained (92\%+) when both training and test data were acquired at the same room temperature. However, the significant drop to 60\% for the highest temperature difference of 17K between training and test data demonstrated the lack of robustness of the proposed features and highlighted the need for the development of discriminative features exhibiting a higher degree of invariance to confounding variables expected in practice. Yoshitomi \textit{et al.}\ did not investigate the effectiveness of their method in the presence of other nuisance factors, such as pose or expression.

\subsubsection{Infrared Local Binary Patterns}\label{sss:lbps}
In a series of influential works, Li \textit{et al.}\ \cite{Li2005,LiChuAoZhan+2006,LiZhanLiaoZhu+2006a,LiChuLiaoZhan2007} were the first to use features based on local binary patterns (LBP) \cite{OjalPietHarw1994} extracted from IR images. They apply their algorithm in an active setting which uses strong NIR light-emitting diodes, coaxial with the direction of the camera. This setup ensures both that the face is illuminated as homogeneously as possible, thus removing the need of algorithmic robustness to NIR illumination, as well as that the eyes can be reliably detected using the bright pupil effect. Evaluated in an indoor setting and with cooperative users, their system achieved impressive accuracy. However, as noted by Li \textit{et al.}\ \cite{LiChuLiaoZhan2007} themselves, their approach was unsuitable for uncooperative user applications or outdoor use due to the strong NIR component of sunlight.

The use of local binary patters was also investigated by Maeng \textit{et al.}\ \cite{MaenChoiParkLee+2011}, who applied them in a multi-scale framework on NIR imagery acquired at distance (up to 60m) with limited success, dense SIFT based features proving more successful in their recognition scenario. Comparative evaluation of local binary patters in the context of a variety of linear and kernel methods was recently published by Goswami \textit{et al.}\ \cite{GoswChanWindKitt2011}.

\subsubsection{Wavelet Transform}\label{sss:wavelet}
Owing to its ability to capture both frequency and spatial information, the wavelet transform has been studied extensively as a means of representing a wide range of 1D and 2D signals, including face appearance in the visual spectrum. Srivastava \textit{et al.}\ \cite{SrivLiuThomHesh2001,SrivLiu2003} were the first to investigate the use of wavelets for extracting robust features from face ``appearance'' images in the IR spectrum. They described a system which uses the wavelet transform based on a bank of Gabor filters. The marginal density functions of the filtered features are then modelled using \emph{Bessel K forms}, which are matched using the simple $L_2$-norm. Srivastava \textit{et al.}\ reported a remarkable fit between the observed and the estimated marginals across a large set of filtered images. Evaluated on the Equinox database their method achieved a nearly perfect recognition rate and on the FSU database outperformed both Eigenfaces and independent component analysis based matching. A similar approach was also described by Buddharaju \textit{et al.}\ \cite{BuddPavlKaka2004}. The method of Nicolo and Schmid \cite{NicoSchm2011} also adopts Gabor wavelet features at its core and encodes the responses using the recently introduced Weber local descriptor \cite{ChenShanHeZhao+2010} and local binary patterns.

\subsubsection{Curvelet Transform}\label{sss:curvelet}
The curvelet transform an extension of the wavelet transform in which the degree of orientational localization is dependent on the scale of the curvelet \cite{MandMajuWu2007}. For a variety of natural images, the curvelet transform facilitates a sparser representation than wavelet transforms do, with effective spatial and directional localization of edge-like structures. Xie \textit{et al.}\ \cite{XieWuLiuFang2009a,XieWuLiuFang2009b,XieLiuWuLu2009c} described the first IR based AFR system which uses the curvelet transform for feature extraction. Using a simple nearest neighbour classifier, in their experiments the method demonstrated a slight advantage (of approximately 1-2\%) over simple linear discriminant based approaches, but with a significant improvement in computational and storage demands.

\subsubsection{Vascular Networks}\label{sss:vascular}
Although the idea of using the superficial vascular network of a face to derive robust features for recognition dates as far back as the work of Prokoski \textit{et al.}\ \cite{ProkRiedCoff1992}, it wasn't until only recently that the first automatic methods have been described in the literature. Following automatic background-foreground segmentation of a face, the method proposed by Buddharaju \textit{et al.}\ \cite{BuddPavlTsiaBaza2007} first extracts blood vessels from an image using simple morphological filters. The skeletonized vascular network is then used to localize salient features of the network which they term \emph{thermal minutia points} and which are similar in nature to the minutiae used in fingerprint recognition. Indeed, the authors adopt a method of matching sets of minutia points already widely used in fingerprint recognition, using relative minutiae orientations on local and global scale (a similar method was subsequently described by Seal \textit{et al.}\ \cite{SealNasiBhatBasu2011}). Unsurprisingly, the method's performance was best when the semi-profile pose was used for training and querying, rather than the frontal pose. This finding is similar to what has repeatedly been noted by multiple authors for both human and computer based recognition in the visible spectrum \cite{SimZhang2004,LeeMoghPfisMach2004,AranCipo2004a}. While images of frontally oriented faces contain the highest degree of appearance redundancy, they limit the amount of discriminative information available from the sides of the face. In the multi-pose training scenario, rank-1 recognition of approximately 86\% and the equal error rate of approximately 18\% were achieved. While, as the authors note, some of the errors can be attributed to incorrectly localized thermal minutia points, the main reason for the relatively poor performance of their method is to be found in the sensitivity of their geometry based approach to out-of-plane rotation and the effected distortion of the observed vascular network shape.

In their more recent work, Buddharaju \textit{et al.}\ \cite{BuddPavl2009} improve their method on several accounts. Firstly, they introduce a post-processing step in their vascular network segmentation algorithm, with the aim of removing spurious segments which, as mentioned previously, are responsible for some of the matching errors observed of their initial method \cite{BuddPavlTsiaBaza2007}. More significantly, using an iterative closest point algorithm Buddharaju \textit{et al.}\ now also non-rigidly register two vascular networks which are being compared as a means of correcting for the distortion effected by out-of-plane head rotation. Their experiments indeed demonstrate the superiority of this approach over that proposed previously.

Cho \textit{et al.}\ \cite{ChoWangOng2009} describe a simple modification of the temporal minutia point based approach of Buddharaju \textit{et al.}\ which appends the location of the face centre (estimated from the segmented foreground mask) to the vectors corresponding to minutia point loci. Their method significantly outperformed Na\"{\i}ve Bayes, multilayer perceptron and Adaboost classifiers, achieving a false acceptance rate of 1.2\% for the false rejection rate of 0.1\% on the Equinox database.

The most recent contribution to the corpus of work on vascular network based recognition was made by Ghiass \textit{et al.}\ \cite{GhiaAranBendMald2013,GhiaAranBendMald2013a}. There are several important aspects of novelty in the approach they describe. Firstly, instead of seeking a binary representation in which each pixel either `crisply' belongs or does not belong to the vascular network, the baseline representation of Ghiass \textit{et al.}\ smoothly encodes this membership by a confidence level in the interval $[0,1]$. This change of paradigm, further embedded within a multi-scale vascular network extraction framework, is shown to achieve better robustness to face scale changes (e.g.\ due to different resolutions of query and training images, or indeed different user-camera distances). The second significant contribution of this work concerns the recognition across pose which is a major challenge for previously proposed vascular network based methods. The method of Ghiass \textit{et al.}\ achieves pose invariance by geometrically warping images to a canonical frame. Ghiass \textit{et al.}\ are the first to show how the active appearance model (AAM) \cite{CootEdwaTayl1998} can be applied on IR images of faces and, specifically, they show how the difficult problem of AAM convergence in the presence of many local minima can be addressed by pre-processing thermal IR images in a manner which emphasizes discriminative information content~\cite{GhiaAranBendMald2013}. In their most recent work, recognition across the entire range of poses from frontal to profile is achieved by training en ensemble of AAMs, each `specializing' in a particular region of the thermal IR face space corresponding to an automatically determined cluster of poses and subject appearances~\cite{GhiaAranBendMald2013a}.

Lastly, it should be noted that Ghiass \textit{et al.}\ emphasize that ``$\ldots$none of the existing publications on face recognition using `vascular network' based representations provide any evidence that the extracted structures are indeed blood vessels. Thus the reader should understand that we use this term for the sake of consistency with previous work, and that we do \emph{not} claim that what we extract in this paper is an actual vascular network. Rather we prefer to think of our representation as a \emph{function} of the underlying vasculature'' (the reader may also find the work of Gault \textit{et al.}\ \cite{GaulBlumFaraStar2010} useful in the consideration of this issue).


\subsubsection{Blood Perfusion}\label{sss:perfusion}
A different attempt at extracting invariant features which also exploits the temperature differential between vascular and non-vascular tissues was proposed by Wu \textit{et al.}\ \cite{WuSongJianXie+2005} and Xie \textit{et al.}\ \cite{XieLiuWuFang2009}. Using a series of assumptions on relative temperatures of body's deep and superficial tissues, and the ambient temperature, Wu \textit{et al.}\ formulate a differential equation governing blood perfusion. The model is then used to compute a ``blood perfusion image'' from the original segmented thermogram of a face. Finally, blood perfusion images are matched using a standard linear discriminant and an RBF network.

Following their original work, Wu \textit{et al.}\ \cite{WuGuChiaOng2007a} and Xie \textit{et al.}\ \cite{XieWuHeFang+2010} introduce alternative blood perfusion models. The model described by Wu \textit{et al.}\ was demonstrated to produce comparable recognition results to the more complex model previously, while achieving greater time and storage efficiency. Xie \textit{et al.}\ derived a model based on the Pennes equation which too outperformed the initial model described by Wu \textit{et al.}\ \cite{WuSongJianXie+2005}.

In addition to their work on different blood perfusion models, in their more recent work Wu \textit{et al.}\ \cite{WuWeiFangLi+2007} also extend their classification method by another feature extraction stage. Instead of using the blood perfusion image directly, they first decompose the image of a face using the wavelet transform. After that, they apply the sub-block discrete cosine transform on the low frequency sub-band of the transform and use the obtained coefficients as an identity descriptor. Wu \textit{et al.}\ demonstrate experimentally that this representation outperforms both purely discrete cosine transform based and purely wavelet transform based representations of the blood perfusion image.

\subsection{Multi-Spectral and Hyper-Spectral Methods}\label{ss:spectral}
Multi-spectral imaging refers to the process of concurrent acquisition of a set of images, each image corresponding to a different band of the electromagnetic spectrum. A familiar example is colour imaging in the visual spectrum which acquires three images that correspond to what the human eyes perceives as red, green, and blue sensations. In general, the number of bands can be much greater and the width of the sub-bands different images correspond to wider or narrower. The terms multi-spectral and hyper-spectral imaging are often used interchangeably, while some authors make the distinction between sets of images acquired in discrete and separated narrow bands (multi-spectral) and sets of images acquired in usually wider but frequency wise contiguous sub-bands. Henceforth in this paper we will consistently use the term multi-spectral imaging and specifically describe the data used by a specific method.

The epidermal and dermal layers of skin make up a scattering medium that contains pigments such as melanin, hemoglobin, bilirubin, and $\beta$-carotene. Small changes in the distribution of these pigments induce significant changes in the skin's spectral reflectance. In the method of Pan \textit{et al.}\ \cite{PanHealPrasTrom2004}, the structure of the skin, including sub-surface layers, is sensed using multi-spectral imaging in 31 narrow bands of the NIR sub-band. The authors measured the variability in spectral properties of the human skin and showed that there are significant differences in both amplitude and spectral shape of the reflectance curves for the different subjects, while the spectral reflectance for the same subject did not change in different trials. They also observed good invariance of local spectral properties to face orientation and expression. On a proprietary database of 200 subjects with a diverse sex, age and ethnicity composition, the proposed method achieved recognition rates of 50\%, 75\%, and 92\% for profile, semi-profile and frontal faces respectively. In their subsequent work, Pan \textit{et al.}\ \cite{PanHealPrasTrom2005} examine the use of holistic multi-spectral appearance, in contrast to their previous work which used a sparse set of local features only. They apply Eigenfaces on images obtained from different NIR sub-bands, as a means of de-correlating the set of features used for classification. They also describe a method for synthesizing a discriminative signature image that they term the ``spectral-face'' image, obtained by sequential interlacing of images corresponding to different sub-bands, which in their experiments showed some advantage when used as input for Eigenfaces.


\subsubsection{Inter-Spectral Matching}\label{sss:interspectral}
The work by Bourlai \textit{et al.}\ \cite{BourKalkRossCuki+2010} is the only published account of the use of data acquired in the SWIR sub-band for AFR. Following face localization using the detector of Viola and Jones \cite{ViolJone2004}, Bourlai \textit{et al.}\ apply contrast limited adaptive histogram equalization and feed the result into: (i) a $K$-nearest neighbour based classifier, (ii) VeriLook's and (iii) Identity Tools G8 commercial recognition systems. A particularly interesting aspect of this work is that Bourlai \textit{et al.}\ investigate the possibility of inter-spectral matching. Their experimental results suggest that SWIR images can be matched to visible images with promising results. Klare and Jain \cite{KlarJain2010} similarly match visible and NIR data, using local binary patterns and HoG local descriptors \cite{DalaTrig2005}. The success of these methods not particularly surprising considering that the NIR and SWIR sub-bands of the IR spectrum is much closer to the visible spectrum than MWIR or LWIR sub-bands. Indeed, this premise is central to the methods described by Chen \textit{et al.}\ \cite{ChenYiYangZhao+2009}, Lei and Li \cite{LeiLi2009}, Mavadati \textit{et al.}\ \cite{MavaSadeKitt2010} and Shao \textit{et al.}\ \cite{ShaoWangWang2009} who show that visible spectrum data can be used to create synthetic NIR images, the NIR sub-band of the IR being the closest to the visible spectrum.

A greater challenge was recently investigated by Bourlai \textit{et al.}\ \cite{BourRossChenHorn2012} who attempted to match MWIR to visible spectrum images. Following global affine normalization and contrast limited adaptive histogram equalization, the authors evaluated different pre-processing methods (the self-quotient image and difference of Gaussian based filtering), feature types (local binary patterns, pyramids of oriented gradients histograms \cite{BoscZissMuno2007} and scale invariant feature transform \cite{Lowe2004}) and similarity measures (chi-squared, distance transform based, Euclidean and city-block). No combination of the parameters was found to be very promising, the best performing patch based and difference of Gaussian filtered LBP on average achieving only approximately 40\% correct rank-1 recognition rate on a 39 subject subset of the West Virginia University database.

\subsection{Multimodal Methods}\label{ss:multimodal}
As predicted from theory and repeatedly demonstrated in experiments summarized in the preceding sections, some of the major challenges of AFR methods which use IR images include the opaqueness of eyeglasses in this spectrum and the dependence of the acquired data on the emotional and physical condition of the subject. In contrast, neither of these is a significant challenge in the visible spectrum. In the visible spectrum eyeglasses are largely transparent and such physiological variables such as the emotional state have negligible inherent effect on one's appearance. Indeed, in the context of many challenging factors in the two spectra, they can be considered complementary. Consequently, it can be expected that this complementary information can be exploited to achieve a greater degree of invariance across a wide range of nuisance variables.

Most of the methods for fusing information from visible and IR spectra described in the literature fall into one of two groups. The first of these is data-level fusion. Methods of this category construct features which inherit information from both modalities, and then perform learning and classification of such features. The second fusion type is decision-level. Methods of this group compute the final score of matching two individuals from matches independently performed in the visible and in the IR spectra. To date, decision-level fusion predominates in the IR AFR literature.

\subsubsection{Early Work}
Wilder \textit{et al.}\ \cite{WildPhilJianWien1996} were the first to investigate the possibility of fusion of visible and IR data. They examined three different methods for representing and matching images, using (i) transform coded grey scale projections, (ii) Eigenfaces and (iii) pursuit filters, and compared the performance of the two modalities in isolation and their fusion. Decision-level fusion was achieved simply by adding the matching scores separately computed for visible and IR data. The transform coded grey scale projections based method achieved the best performance of the three methods compared. Using this representation independently in the visible and IR spectra, the two modalities achieved comparable recognition results. However, the proposed fusion method had a remarkable effect, reducing the error rate for approximately an order of magnitude (from approximately 10\% down to approximately 1\%).


\subsubsection{Time-Lapse}
The problem of time-lapse in recognition concerns the empirical observation made across different recognition methodologies that the performance of an algorithm degrades with the passage of time between training and test data even if the acquisition conditions are seemingly the same. The term ``time-lapse'' is, we would argue, a somewhat misleading one. Clearly, the drop in recognition performance is not caused by the passage of time \textit{per se} but rather a change in some tangible factor which affects facial appearance. This is particularly easy to illustrate on thermal data. Even if external imaging conditions are controlled or compensated for, none of the published work attempts to control or measure the effects of the emotional state or the level of excitement of the subject or indeed the loss of calibration of the IR camera \cite{Mald2001}. The effects of external temperature on the temperature of the face is explicitly handled only in the method proposed by Siddiqui \textit{et al.}\ \cite{SiddSherKhal2004} who used simple thresholding and image enhancement to detect and normalize the appearance of face regions with particularly delayed temperature regulation. Nonetheless, for the sake of consistency and uniformity with the rest of the literature, we shall continue using the term ``time-lapse'' with an implicit understanding of the underlying issues raised herein.

The effect of time-lapse on the performance of IR based systems was investigated by Chen \textit{et al.}\ \cite{ChenFlynBowy2003}. They presented experiments evidencing the complementarity of visible and IR spectra in the presence of time-lapse by showing that recognition errors achieved using the two modalities, and effected by the passage of time between training and query data acquisition, are largely uncorrelated. Similar observations were made by Socolinsky \textit{et al.}\ \cite{SocoSeli2004}. Regardless of whether simple PCA features were used for matching or the commercial system developed by the Equinox Corporation, the benefit of fusing visible and IR modalities was substantial even though the simple additive combination of matching scores was used.

\subsubsection{Eyeglasses}\label{sss:glasses}
Since eyeglasses are opaque to the IR frequencies in the SWIR, MWIR and LWIR sub-bands \cite{TasmJaeg2009}, their presence is a major issue when this data is used for recognition as some of the most discriminative regions of the face can be occluded. In contrast, the effect of eyeglasses on the appearance in the visible spectrum is far less significant. The methods of Gyaourova \textit{et al.}\ \cite{GyaoBebiPavl2004} and Singh \textit{et al.}\ \cite{SingGyaoBebi+2004} propose a data level fusion approach whereby a genetic algorithm is used to select features computed separately in the visual and IR spectra. Using two types of features, Haar wavelet based and eigencomponent based, and the Equinox database the proposed fusion method was shown to yield a superior performance compared to both purely visual and purely IR based matching, and particularly so in the presence of eyeglasses or variable illumination. Chen \textit{et al.}\ \cite{ChenJingXiao2005a} describe a similar fusion method. Instead of a genetic algorithm, they employ a fuzzy integral neural network based feature selection algorithm which has the advantage of faster convergence and greater probability of reaching a solution close to the global optimum.

Heo \textit{et al.}\ \cite{HeoKongAbidAbid2004} investigate both data-level and decision-level fusion. First, following the detection of eyeglasses, the corresponding image region is replaced with a generic eye template. As expected, the replacement of the eyeglass region with a generic template significantly improves recognition in the thermal but not in the visible spectrum. Data-level fusion is achieved by simple weighted addition of the corresponding pixels in mutually co-registered visible and IR images. The key contribution of this work pertains to the difference in performance observed between data-level and decision-level fusion. Interestingly, unlike in the case of data-level fusion where a remarkable performance improvement was observed, when fusion was performed at the decision-level the performance was actually somewhat worsened.

A similar approach to handing the occlusion of IR image regions by eyeglasses was taken by Kong \textit{et al.}\ \cite{KongHeoBougZheng+2007}. They replace an elliptical patch surrounding the eye occluded by eyeglasses with a patch representing the average eye appearance. Although differently implemented, the approach of Arandjelovi\'c \textit{et al.}\ \cite{AranHammCipo2006,AranHammCipo2010} is similar in spirit. Following the detection of eyeglasses unlike Heo \textit{et al.}\ and Kong \textit{et al.}\ Arandjelovi\'c \textit{et al.}\ do not remove the offending image region, but rather introduce a robust modification to canonical correlations based matching, which ignores the eyeglasses region when sets of images are compared.

\subsubsection{Illumination}
In addition to the problem posed by eyeglasses, in their work already described in Sec.~\ref{sss:glasses} Heo \textit{et al.}\ \cite{HeoKongAbidAbid2004} also examined the effects of the proposed fusion on illumination invariance. Their results successfully substantiated the theoretically expected complementarity of IR and visible spectrum data. Socolinsky \textit{et al.}\ \cite{SocoSeli2004} extend their previous work \cite{SocoSeli2002} by describing a simple decision based fusion based on a weighted combination of visible and IR based matching scores, and evaluate it in indoor and outdoor data acquisition environments. The more extreme illumination conditions encountered outdoors proved rather more challenging than the indoor environment, regardless of which modality or baseline matching algorithm was used for recognition. Although simple, their fusion approach did yield substantial improvements in all cases, but still failed reach practically useful performance levels when applied outdoors. Bhowmik \textit{et al.}\ \cite{BhowBhatNasiBasu+2010} also investigated a simple weighted combination of visible and IR spectrum matching scores.

The limitations of the approaches of Socolinsky \textit{et al.}\ and Bhowmik \textit{et al.}\ was recognized by Arandjelovi\'c \textit{et al.}\ \cite{AranHammCipo2006a,AranHammCipo2010}, who demonstrate that the optimal weights in decision-level fusion are illumination dependent. Their main contribution is a fusion method which learns the optimal weighting of matching scores in an illumination specific manner~\cite{AranCipo2006a}. Illumination specificity is achieved implicitly by exploiting the observation that if the best match in the visible domain is sufficiently confident, the illumination change between training and novel data is small so more weight should be placed on the visible spectrum match and \textit{vice versa}. Conceptually similar is the fusion approach described by Moon \textit{et al.}\ \cite{MoonKongYooChun2006} which also adaptively controls the contributions of the visible and IR spectra. Unlike Arandjelovi\'c \textit{et al.}\ who use a combination of filtered holistic and local appearances, Moon \textit{et al.}\ represent images of faces using the coefficients obtained from a wavelet decomposition of an input image. Different wavelet based fusion approaches have also been proposed by Kwon \textit{et al.}\ \cite{KwonKong2005} and Zahran \textit{et al.}\ \cite{ZahrAbbaDessAsho+2009}.

\subsubsection{Expression}
The method proposed by Hariharan \textit{et al.}\ \cite{HariKoscAbidGrib+2006} is one of the small number data-level fusion approaches. Hariharan \textit{et al.}\ produce a synthetic image which contains information from both visible and IR spectra. The key element of their approach is empirical mode decomposition. After decomposing the corresponding and mutually co-registered visible and IR spectrum images into their intrinsic mode functions, a new image is produced as a re-weighted sum of the intrinsic mode functions of both modalities. The re-weighting coefficients are determined experimentally on a training set in an \textit{ad hoc} subjective manner which involves human judgement on how discriminative the resulting image appears. Hariharan \textit{et al.}\ report that their method outperformed that proposed by Kong \textit{et al.}\ \cite{KongHeoBougZheng+2007}, as well as Rockinger and Fechner \cite{RockFech1998}, and particularly so in poor illumination conditions and in the presence of facial expression changes.

\subsection{Other Approaches}
Owing to the increasing popularity of research into IR based recognition there are a number of approaches in the literature which we did not discuss explicitly. These include the geometric invariant moment based approaches of Abas and Ono \cite{AbaOno2010}, elastic graph matching based method of Hizem textit{et al.}\ \cite{HizeAllaMellDori2009}, isotherm based method of Tzeng \textit{et al.}\ \cite{TzenLeeChen2011}, faceprints of Akhloufi and Bendada \cite{AkhlBend2008}, fusion work of Toh \textit{et al.}\ \cite{Toh2010}, and others \cite{ShaoWang2009,ZahrAbbaDessAsho+2009a,PopGordFlorVlai2010}. Specifically, we did not describe (i) minor extensions of the original approaches already surveyed and (ii) those methods which lack the weight of sufficient empirical evidence to support their competitiveness with the state-of-the-art at the time when they were first proposed. Nonetheless references to these are provided herein for the sake of completeness and for the benefit of the reader.


\section{Summary and Conclusions}\label{s:summary}
The use of IR imaging for AFR, as an alternative to visual spectrum based approaches, has attracted substantial research and commercial attention as a modality which could facilitate greater robustness to illumination and facial expression changes, facial disguises and dark environments. In this paper we reviewed a large and rapidly expanding corpus of work in this area. A notable limitation which we found in all of the reviewed publications, is of a methodological nature: despite the universal acknowledgment of the major challenges of IR based AFR, none of the reported experiments examine the proposed methods's performance in the context of all of them. As such, direct comparison of different approaches is difficult as is the assessment of their capacity for practical deployment. Nonetheless, in the opinion of these authors, two particularly promising ideas stand out: (i) the development of identity descriptors based on persistent physiological features (e.g.\ vascular networks), and (ii) the use of methods for multi-modal fusion of complementary data types (e.g.\ visible and IR). Both research directions are still in their early stages, with a potential for significant further improvement.

\bibliographystyle{ieeetran}
\bibliography{./eccv02,./my_bibliography}

\end{document}